\documentclass[10pt,twocolumn,letterpaper]{article}

\usepackage{ifpdf}
\ifpdf
  \pdfoutput=1
\fi

\usepackage{iccv}              
\usepackage{times}
\usepackage{epsfig}
\usepackage{graphicx}
\usepackage{amsmath}
\usepackage{amssymb}
\usepackage{tabularx}
\usepackage{makecell} 
\usepackage{array}    
\usepackage{multirow} 
\usepackage{booktabs}

\PassOptionsToPackage{table,dvipsnames}{xcolor}
\usepackage{xcolor}
\usepackage{pifont}
\usepackage{colortbl}

\definecolor{iccvblue}{rgb}{0.21,0.49,0.74}
\usepackage[pagebackref,breaklinks,colorlinks,allcolors=iccvblue]{hyperref}

\def\paperID{10374} 
\def\confName{ICCV}
\def\confYear{2025}

\title{NuPlanQA: A Large-Scale Dataset and Benchmark for Multi-View\\Driving Scene Understanding in Multi-Modal Large Language Models}

\author{
Sung-Yeon Park\textsuperscript{1}, Can Cui\textsuperscript{1}, Yunsheng Ma\textsuperscript{1},\\Ahmadreza Moradipari\textsuperscript{2}, Rohit Gupta\textsuperscript{2}, Kyungtae Han\textsuperscript{2}, and Ziran Wang\textsuperscript{1}\\
\vspace{-3mm}
\\
\textsuperscript{1}Purdue University, \textsuperscript{2}Toyota InfoTech Labs\\
{\tt\small \{sungyeon, ziran\}@purdue.edu}
}

\begin{document}
\maketitle

\def\colourcheck#1{%
  \textcolor{#1}{\ding{52}}%
}
\def\redcross{{\color{red} \ding{55}}}

\begin{figure*}[!t]
    \centering
    \includegraphics[width=1.0\textwidth]{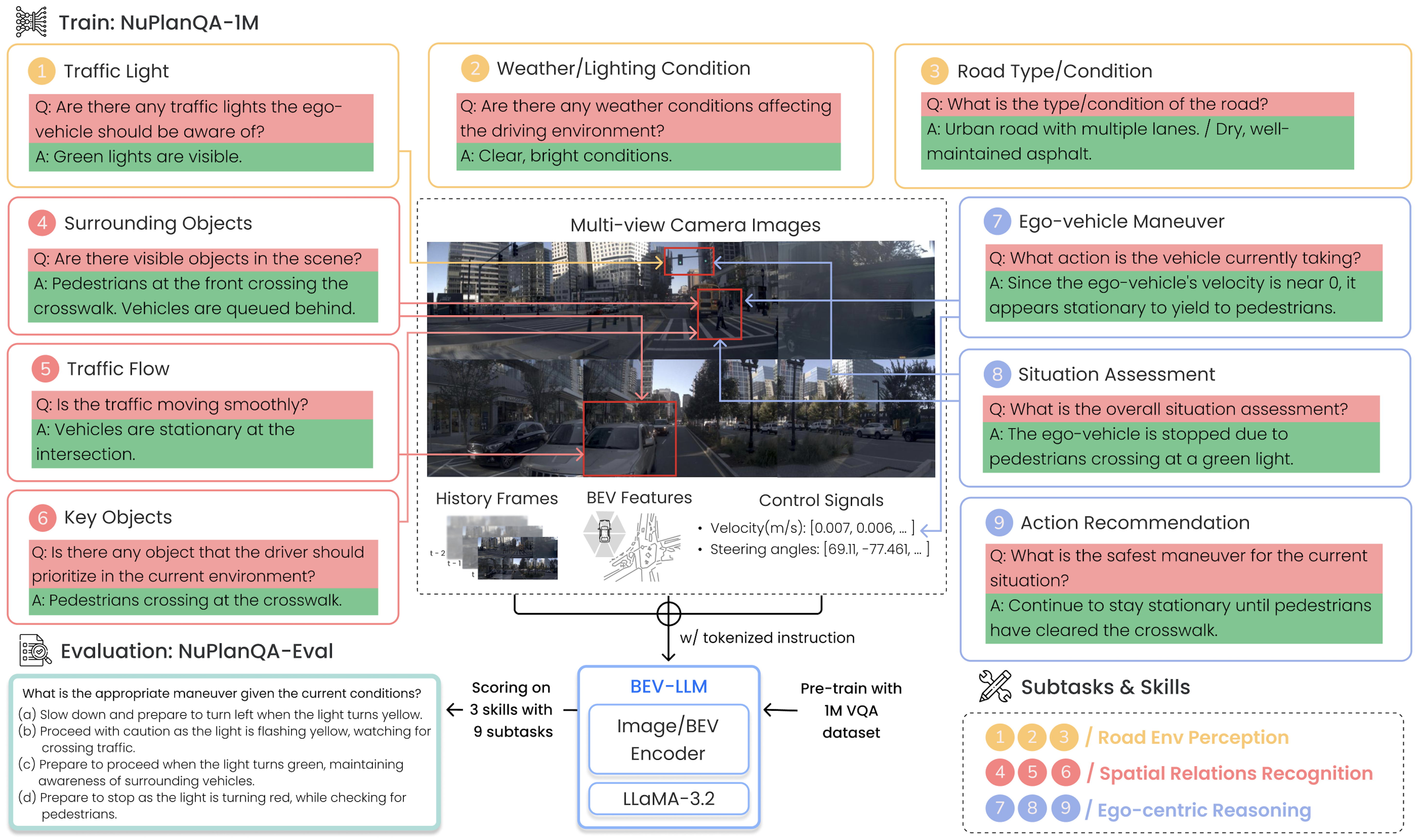}
\caption{\textbf{An overview of NuPlanQA.} NuPlanQA comprises nine subtasks across three skill areas to support the context-aware analysis of traffic scenes. The proposed baseline, \textit{\textbf{BEV-LLM}}, is trained on \textit{\textbf{NuPlanQA-1M}} using historical frames, BEV features from multi-view images, and control signals as inputs. Finally, we evaluate MLLMs using \textit{\textbf{NuPlanQA-Eval}}, a multiple-choice QA benchmark for driving scene understanding.}
    \vspace{-5mm}
    \label{fig:main}
\end{figure*}
\label{sec:intro}

\begin{abstract}
Recent advances in multi-modal large language models (MLLMs) have demonstrated strong performance across various domains; however, their ability to comprehend driving scenes remains less proven. The complexity of driving scenarios, which includes multi-view information, poses significant challenges for existing MLLMs. In this paper, we introduce NuPlanQA-Eval, a multi-view, multi-modal evaluation benchmark for driving scene understanding. To further support generalization to multi-view driving scenarios, we also propose NuPlanQA-1M, a large-scale dataset comprising 1M real-world visual question-answering (VQA) pairs. For context-aware analysis of traffic scenes, we categorize our dataset into nine subtasks across three core skills: Road Environment Perception, Spatial Relations Recognition, and Ego-Centric Reasoning. Furthermore, we present BEV-LLM, integrating Bird's-Eye-View (BEV) features from multi-view images into MLLMs. Our evaluation results reveal key challenges that existing MLLMs face in driving scene-specific perception and spatial reasoning from ego-centric perspectives. In contrast, BEV-LLM demonstrates remarkable adaptability to this domain, outperforming other models in six of the nine subtasks. These findings highlight how BEV integration enhances multi-view MLLMs while also identifying key areas that require further refinement for effective adaptation to driving scenes. To facilitate further research, we publicly release NuPlanQA at \href{https://github.com/sungyeonparkk/NuPlanQA}{\texttt{github.com/sungyeonparkk/NuPlanQA}}.
\end{abstract}    
\section{Introduction}

The advent of multi-modal large language models (MLLMs) has brought significant benefits to autonomous driving by enhancing generalization and interpretability, which are essential for understanding complex real-world scenes \cite{wen2023roadgpt4visionearlyexplorations, chen2024asynchronouslargelanguagemodel, VLP_Pan_2024_CVPR}. Although autonomous vehicles are progressing rapidly, they still require human intervention in challenging scenarios, such as adverse weather conditions, poor lighting, or interactions with unpredictable road agents \cite{corner_cases_Bogdoll_2021, Zhang_2023}. Additionally, the lack of interpretability in their decision-making process hinders the adoption of end-to-end models \cite{hu2023planningorientedautonomousdriving}. To address these challenges, MLLMs have been integrated into autonomous driving systems, enhancing situational awareness, improving reasoning in ambiguous conditions, and providing more interpretable decision-making \cite{cui2023drivespeakenablinghumanlike, you2024v2xvlmendtoendv2xcooperative, chatscene}. By generating human-readable responses, MLLMs have further improved performance across various tasks, seamlessly integrating into pipelines \cite{xu2024drivegpt4interpretableendtoendautonomous, sima2024drivelmdrivinggraphvisual, wang2024omnidriveholisticllmagentframework, wei2024occllamaoccupancylanguageactiongenerativeworld}. 
In this context, traffic scene-specific MLLMs have become increasingly vital for generating accurate and context-aware information, surpassing the capabilities of general-purpose MLLMs. In particular, visual question-answering (VQA) task has been explored as a replacement for various modules, such as motion prediction and object detection, while also enhancing MLLMs' understanding of traffic scenes \cite{zhang2024minidriveefficientvisionlanguagemodels, xu2024drivegpt4interpretableendtoendautonomous, sima2024drivelmdrivinggraphvisual, marcu2024lingoqavisualquestionanswering, ma2023dolphinsmultimodallanguagemodel, gopalkrishnan2024multiframelightweightefficient}. 



Recognizing this potential, there has been a growing demand for VQA datasets focused on driving scenes. Researchers have addressed this by constructing such datasets using existing resources like nuScenes and BDD-X \cite{nuscenes_Caesar_2020_CVPR, bddx_kim2018textual, sima2024drivelmdrivinggraphvisual, nuscenesqa, holistic_nuinstruct, ma2023dolphinsmultimodallanguagemodel}. However, the heavy reliance on these resources limits the diversity of scenarios, which in turn affects the adaptability of MLLMs. Furthermore, while strictly annotated datasets from well-established sources facilitate the cost-effective generation of VQA datasets, they primarily rely on instance-level information, such as object bounding boxes and coordinates. This results in limited QA coverage, with a strong focus on instance-level perception and prediction while overlooking crucial road environment details and broader scene comprehension, as these elements are absent from raw datasets. Additionally, this constraint restricts QAs to a fixed format rather than allowing for rich, free-form responses, reducing the diversity of reasoning-based answers.

To overcome these challenges, some studies have introduced a wide range of free-form VQA datasets by leveraging new scenarios from YouTube videos or self-collected vehicle data \cite{marcu2024lingoqavisualquestionanswering, yang2024genad}. However, these datasets consist solely of front-view images, which greatly limits the availability of surrounding information. Furthermore, these datasets are generally small, which makes them insufficient for effectively adapting MLLMs to diverse traffic conditions \cite{nuscenesqa, xu2024drivegpt4interpretableendtoendautonomous, sima2024drivelmdrivinggraphvisual}. This limitation is particularly critical since most MLLMs used for driving scene understanding are pretrained on web-scale datasets, highlighting the necessity of training on large-scale real-world driving scenarios to mitigate biases. Given the challenges of acquiring massive real-world driving data, some studies have turned to simulated environments \cite{ma_lampilot_2024, sima2024drivelmdrivinggraphvisual}. However, a notable gap still exists between simulated and real-world traffic scenes, making it challenging to extend models to real-world scenarios. \cite{holistic_nuinstruct, paul2024legodrivelanguageenhancedgoalorientedclosedloop}.

Moreover, traffic environments demand a more structured and context-aware analysis to ensure safe and effective planning. For example, a model may successfully detect traffic lights but might struggle to identify which signal is relevant to the ego vehicle. Similarly, it may misinterpret traffic flow, failing to distinguish whether vehicles are stopped due to a red light or waiting to make a turn. Although some existing works categorize VQA pairs into broad skill groups—such as perception, prediction, and planning tasks \cite{sima2024drivelmdrivinggraphvisual, marcu2024lingoqavisualquestionanswering}—more fine-grained categorization is necessary to identify MLLMs' major weaknesses and facilitate targeted skill enhancement.

In addition to the limited availability of VQA datasets that meet these standards, a reliable and easily applicable evaluation benchmark for MLLM-integrated driving models has yet to be established. While some downstream tasks can be effectively evaluated using existing metrics \cite{VLP_Pan_2024_CVPR, zhou2024embodiedunderstandingdrivingscenarios}, text-based metrics such as BLEU \cite{bleu}, METEOR \cite{banerjee-lavie-2005-meteor}, and ROUGE-L \cite{lin-2004-rouge} are insufficient for VQA tasks in traffic scene analysis. These metrics, being inherently text-focused, fail to capture the complexities of traffic scenarios, including temporal and spatial dynamics as well as motion-related details. Furthermore, they often overlook critical aspects such as road rules and causal reasoning, making them unable to differentiate between semantically similar but contextually incorrect responses. Although numerous benchmarks exist for evaluating MLLMs across various domains, a dedicated benchmark for driving scene understanding remains largely unexplored \cite{fu2024videommefirstevercomprehensiveevaluation, yue2023mmmu, pătrăucean2023perceptiontestdiagnosticbenchmark, Ye2023mPLUGOwlME, mmbench}. 

\begin{figure*}[!t]
    \centering
    \includegraphics[width=1.0\textwidth]{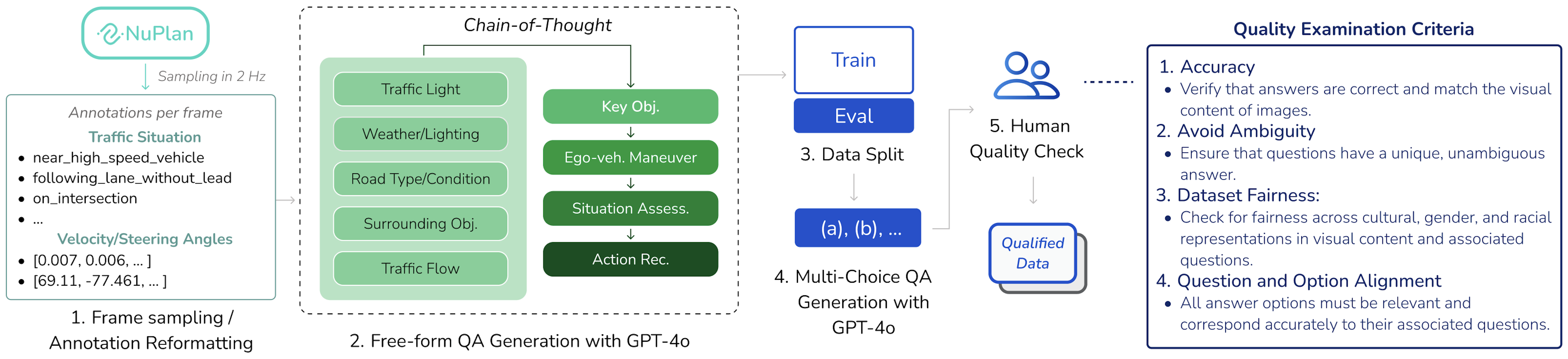}
    \caption{\textbf{Data construction process with human quality checks.} Based on sampled frames and per-frame annotations, free-form QAs are generated by GPT-4o using a chain-of-thought approach. The data is then split into 1M training and 8K evaluation samples, with the evaluation set restructured into multiple-choice QAs. After a quality check following predefined evaluation criteria, the final evaluation dataset is obtained.}
    \vspace{-5mm}
    \label{fig:data-const}
\end{figure*}

Facing these limitations, we develop a large-scale training dataset and an evaluation benchmark grounded in a contextual and systematic analysis of traffic scenes, as well as the baseline model for multi-view VQA. Our three main contributions are as follows:

\noindent {\textbf{1) NuPlanQA Dataset.} We categorize traffic scenes into nine essential elements, each reflecting critical contextual factors: \textit{Traffic Light, Weather/Lighting Condition, Road Type/Condition, Surrounding Objects, Traffic Flow, Key Objects, Ego-Vehicle Maneuver, Situation Assessment, and Action Recommendation}. Using nuPlan \cite{caesar2022nuplanclosedloopmlbasedplanning} and its annotations, we generate free-form QA pairs based on these predefined tasks. Our \textbf{\textit{NuPlanQA-1M}} comprises 1M QA pairs, serving as a large-scale real-world dataset for driving video QA tasks, with all samples featuring multi-view images.

\noindent \textbf{2) Evaluation Benchmark.} 
To establish a universal evaluation standard, we meticulously construct an 8K multiple-choice QA benchmark, \textit{\textbf{NuPlanQA-Eval}}. This benchmark offers a comprehensive assessment of MLLMs in driving scenarios, spanning nine subtasks categorized into three core skills: \textit{Road Environment Perception, Spatial Relations Recognition, and Ego-Centric Reasoning}. Our evaluation results further reveal the key challenges of existing MLLMs through detailed per-task performance analysis.

\noindent \textbf{3) Baseline.} While MLLMs can process both images and videos, they are not inherently designed to handle multi-view image inputs. To bridge this gap, we introduce \textbf{\textit{BEV-LLM}}—a model that integrates BEV features with visual and language modalities, enriching MLLMs with spatially aware BEV representations. As a baseline for our benchmark, BEV-LLM exhibits enhanced reasoning and perception capabilities in complex driving environments.


\section{Methodology}

\subsection{Overview}

To enhance MLLMs' ability to comprehend traffic scenes based on key factors, we categorize our dataset into nine subtasks, ranging from low-level perception to high-level reasoning. As illustrated in Figure \ref{fig:main}, the framework begins with the detection of \textit{(1) traffic light, (2) weather/lighting condition, and (3) road type/condition}. These tasks focus on static visual cues, which contribute to the development of the \textit{\textbf{Road Environment Perception}} skill.
Beyond static features, the ego vehicle must also comprehend spatial relationships between road objects. To address this, we design the following subtasks: \textit{(4) surrounding objects, (5) traffic flow, and (6) key objects}, which collectively develop the model's \textbf{\textit{Spatial Relations Recognition}} skill.
Finally, to further refine MLLMs’ ability to interpret scenes from an ego-centric perspective—as if it were driving inside a vehicle—we introduce the final set of subtasks under the \textbf{\textit{Ego-Centric Reasoning}} category. These include \textit{(7) ego-vehicle maneuver}, which focuses on understanding vehicle control parameters, \textit{(8) situation assessment}, which involves evaluating the scene using the previously gathered information, and \textit{(9) action recommendation}, which suggests appropriate future maneuvers prioritizing safety. This structured, hierarchical approach to scene understanding not only improves the quality of QA dataset generation (as detailed in the following section) but also facilitates a more targeted skill enhancement for the model. Moreover, categorizing tasks at different levels allows for more granular comparisons between MLLMs. Detailed descriptions and examples of each subtask are provided in the supplementary material.

\begin{figure}[!t]
    \centering
    \includegraphics[width=\linewidth]{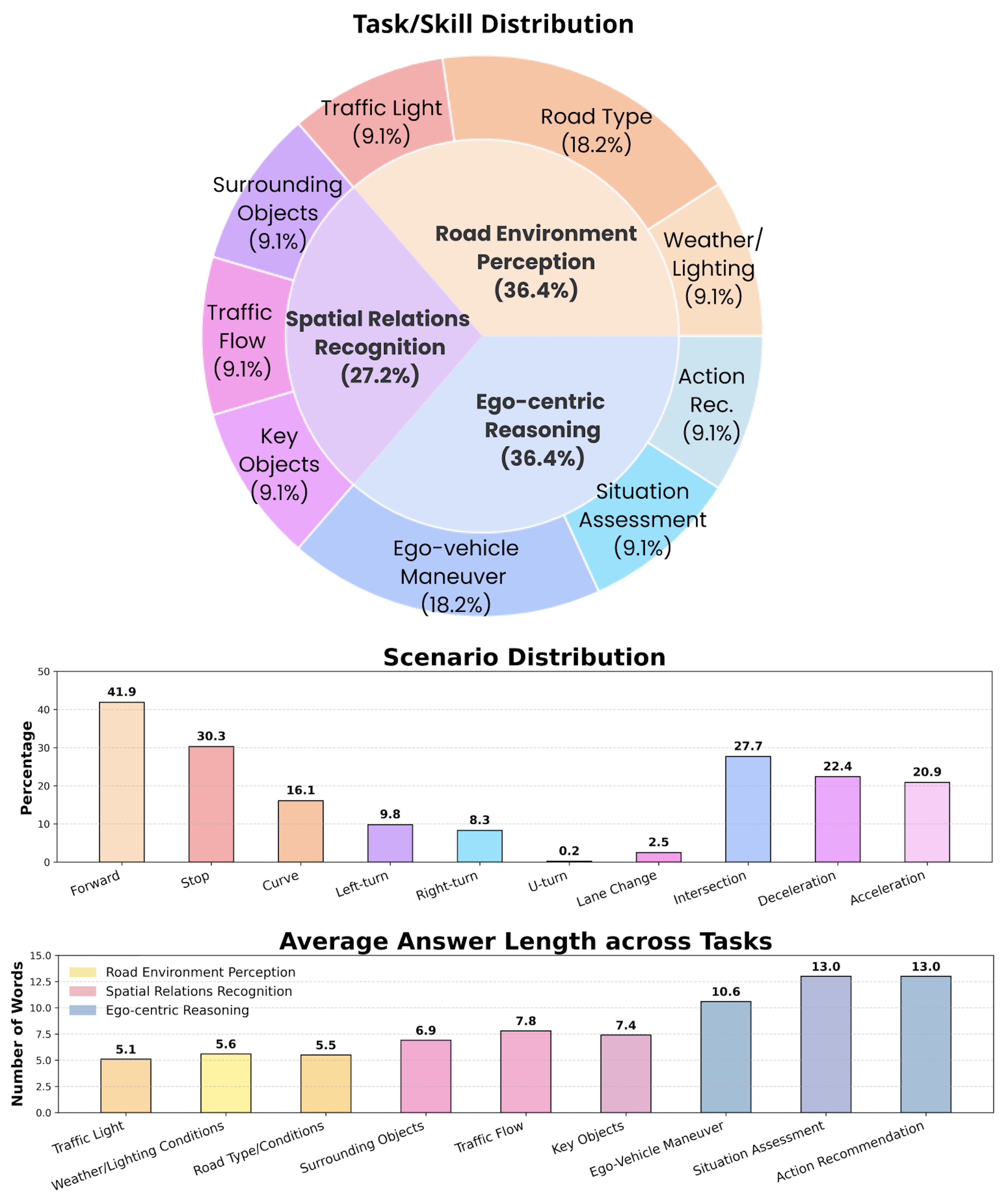}
    \caption{\textbf{Distribution of NuPlanQA-1M.} NuPlanQA-1M maintains a balanced distribution across subtasks and skills. It also covers a wide range of scenarios, with answer length increasing according to the level of tasks.}
    \vspace{-5mm}
    \label{fig:data-stat}
\end{figure}

\begin{table*}[!t]
\centering
  \small
    \raggedright
    \fboxsep=0pt 
    \begin{tabular}{>{\hspace{0pt}}l | c c c c | c c c c c c c c>{\hspace{0pt}}c}  
    \toprule
    Dataset&\#Frames&\#QA&\makecell{Multi\\View}&\makecell{\%Free\\Form}&\makecell{Trfc.\\Light} & \makecell{Wea\\-ther} & \makecell{Road\\Type} 
    & \makecell{Sur.\\Obj.} & \makecell{Trfc.\\Flow} & \makecell{Key\\Obj.} & \makecell{Ego\\Ctrl.} & \makecell{Situ.\\Asse.} & \makecell{Act.\\Rec.}\\
    \midrule
    \textsuperscript{\dag}NuScenes-QA \cite{nuscenesqa}&34K&460K&\colourcheck{green}&0&\redcross&\redcross&\redcross&\colourcheck{green}&\redcross&\redcross&\redcross&\redcross&\redcross\\
    
    \textsuperscript{\dag}NuInstruct \cite{holistic_nuinstruct}&12K&91K&\colourcheck{green}&8&\redcross&\redcross&\redcross&\colourcheck{green}&\redcross&\redcross&\colourcheck{green}&\redcross&\colourcheck{green}\\
    
    \textsuperscript{\dag}NuPrompt \cite{nuprompt}&34K&35K&\colourcheck{green}&0&\redcross&\redcross&\redcross&\colourcheck{green}&\redcross&\redcross&\redcross&\redcross&\redcross\\
    
    \textsuperscript{\dag}DriveLM \cite{sima2024drivelmdrivinggraphvisual}&5K&443K &\colourcheck{green}&20&\colourcheck{green}&\colourcheck{green}&\redcross&\colourcheck{green}&\redcross&\colourcheck{green}&\redcross&\redcross&\colourcheck{green}\\
    
    LingoQA \cite{marcu2024lingoqavisualquestionanswering}&28K&420K&\redcross&100&\colourcheck{green}&\colourcheck{green}&\redcross&\colourcheck{green}&\colourcheck{green}&\colourcheck{green}&\colourcheck{green}&\colourcheck{green}&\colourcheck{green}\\

    BDD-X \cite{bddx_kim2018textual}&7K*&26K&\redcross&100&\colourcheck{green}&\colourcheck{green}&\redcross&\colourcheck{green}&\redcross&\colourcheck{green}&\colourcheck{green}&\colourcheck{green}&\redcross\\

    DRAMA \cite{malla2022dramajointrisklocalization}&17K*&17K&\redcross&100&\colourcheck{green}&\redcross&\redcross&\colourcheck{green}&\redcross&\colourcheck{green}&\redcross&\colourcheck{green}&\colourcheck{green}\\
    
    VLAAD \cite{vlaad}&13K*&64K&\redcross&100&\colourcheck{green}&\colourcheck{green}&\redcross&\colourcheck{green}&\colourcheck{green}&\colourcheck{green}&\colourcheck{green}&\colourcheck{green}&\colourcheck{green}\\
    \midrule
    
\rowcolor{gray!25}\textbf{NuPlanQA-1M}{\scriptsize(eval)}&\textbf{90K}{\scriptsize(4K)}&\textbf{1M}{\scriptsize(8K)}&\colourcheck{green}&90\rule{0pt}{1.0em}&\colourcheck{green}&\colourcheck{green}&\colourcheck{green}&\colourcheck{green}&\colourcheck{green}&\colourcheck{green}&\colourcheck{green}&\colourcheck{green}&\colourcheck{green}\\
    
    \bottomrule
    \end{tabular}
    \caption{\textbf{Comparison between NuPlanQA and existing VQA datasets for driving scene understanding.} Our proposed dataset surpasses existing datasets in the scale of annotated front-view frames, the number of QAs, the proportion of free-form QAs, and the inclusion of multi-view images. It also encompasses a wider range of tasks. \textsuperscript{\dag} indicates a nuScenes-based dataset. * represents annotated videos. {\scriptsize(small text)} indicates NuPlanQA-Eval. The proportion of free-form QAs is calculated, excluding short-answer QAs.}
    \vspace{-5mm}
    \label{tab:dataset-comparison}
\end{table*}

\subsection{NuPlanQA}

Our dataset is built upon nuPlan \cite{caesar2022nuplanclosedloopmlbasedplanning}, leveraging its multi-view images, sensor configurations, and high-quality annotations. The selected portion of nuPlan consists of 4.3M frames collected at 10 Hz from four cities—Boston, Pittsburgh, Las Vegas, and Singapore—offering a comprehensive resource for training and evaluation.

\vspace{.5em} 

\noindent \textbf{NuPlanQA-1M.} 
Figure \ref{fig:data-const} illustrates the construction process of our proposed dataset, where frames are sampled at approximately 2 Hz, and annotations are reformatted with descriptions of traffic situations, velocity, and steering wheel angles. These annotations enable GPT-4o \cite{openai2024gpt4ocard} to infer the ego vehicle's current maneuver, such as making a right turn or accelerating, while also capturing scene context from descriptions like ``following lane without lead". With these annotations, multi-view images and past frames spanning 1.5 seconds are fed into GPT-4o to generate QA pairs. Given the GPT-4o’s inherent variability, chain-of-thought prompting \cite{wei2023chainofthoughtpromptingelicitsreasoning} is applied to enhance consistency. The hierarchical structure of the subtasks is preserved during data generation, ensuring that high-level reasoning tasks build upon previously revealed environmental and spatial contexts. The prompts used for this process are provided in the supplementary material. After partitioning the generated pairs into training and evaluation sets, approximately 1M QA pairs are allocated to the training set. The distribution of each subtask, scenario type, and answer length is shown in Figure \ref{fig:data-stat}. This distribution reflects a balanced representation of subtasks and diverse driving scenarios. Additionally, answer length tends to increase for tasks that require more descriptive and reasoning-based responses, aligning with the hierarchical nature of the dataset. 

\vspace{.5em} 

\noindent \textbf{NuPlanQA-Eval.} While most MLLM benchmarks employ multiple-choice QA formats for robust evaluation \cite{pătrăucean2023perceptiontestdiagnosticbenchmark, fu2024videommefirstevercomprehensiveevaluation}, research on MLLMs in autonomous driving often designs custom tasks such as object localization or tracking, which lack universal applicability for MLLM evaluation. To address this, our benchmark adopts a multiple-choice QA format, ensuring both robustness and broader applicability. After generating NuPlanQA-1M, a subset of pre-generated QA pairs is selected for evaluation, with false answer options augmented using GPT-4o. To uphold dataset quality, human annotators manually review all QA pairs, following the quality assessment criteria outlined on the right side of Figure \ref{fig:data-const}. Whereas our proposed NuPlanQA-1M has a higher proportion of road type and ego-vehicle maneuver tasks, NuPlanQA-Eval maintains an almost equal distribution across all tasks to ensure fair evaluation. Further details on dataset statistics are provided in the supplementary material. The final evaluation benchmark is structured into three subsets: a training set (4,634 pairs, $\sim$57\%) for fine-tuning or few-shot prompting, a validation set (1,750 pairs, $\sim$21\%), and a held-out test set (1,801 pairs, $\sim$22\%).

\subsection{Dataset Comparison}
In this part, we compare NuPlanQA with existing datasets for the VQA task in driving scenes. As shown in Table \ref{tab:dataset-comparison}, NuPlanQA-1M stands out by providing 1M QA pairs, significantly surpassing other datasets in scale. In addition to the number of QA pairs, our dataset includes 90K front-view frames, more than twice the amount found in other datasets. Beyond scale, the structure and quality of QA pairs are equally crucial in enhancing model performance. Many nuScenes-based datasets, however, primarily focus on fixed-form grounded VQA, where responses are explicitly tied to specific objects within the image. This results in a strong bias toward instance-level perception, limiting their ability to reason about the broader scene. Their restricted task coverage, as shown on the right side of Table \ref{tab:dataset-comparison}, further stems from the inherent limitations of nuScenes, which does not provide the necessary data for extracting such tasks. Additionally, the prevalence of binary and short-answer QAs, coupled with the focus on grounded VQA, constrains MLLMs' ability to fully understand and describe scenes due to the limited corpus size. Moreover, the rigid phrasing of QA pairs in nuScenes-based datasets, as illustrated in Figure \ref{fig:qa-task}, further limits their expressiveness, reducing fluency and naturalness in scene descriptions.

While datasets such as LingoQA \cite{marcu2024lingoqavisualquestionanswering}, BDD-X \cite{bddx_kim2018textual}, DRAMA \cite{malla2022dramajointrisklocalization}, and VLAAD \cite{vlaad} offer richer textual descriptions and cover diverse tasks, they are limited to front-view images. Although front-view images provide essential information for forward motion and object detection, multi-view images significantly enhance perception, tracking, depth estimation, and occlusion handling. Furthermore, a closer examination of these datasets reveals that, despite their task diversity, they do not consistently cover critical elements such as traffic lights, weather/lighting, road types, and traffic flow across all frames or videos. To overcome these limitations, NuPlanQA introduces over 90\% free-form QA pairs (excluding short-answer) with broad task coverage across all frames, allowing for more comprehensive and flexible scene descriptions.


\begin{figure}[t]
    \centering
    \includegraphics[width=\linewidth]{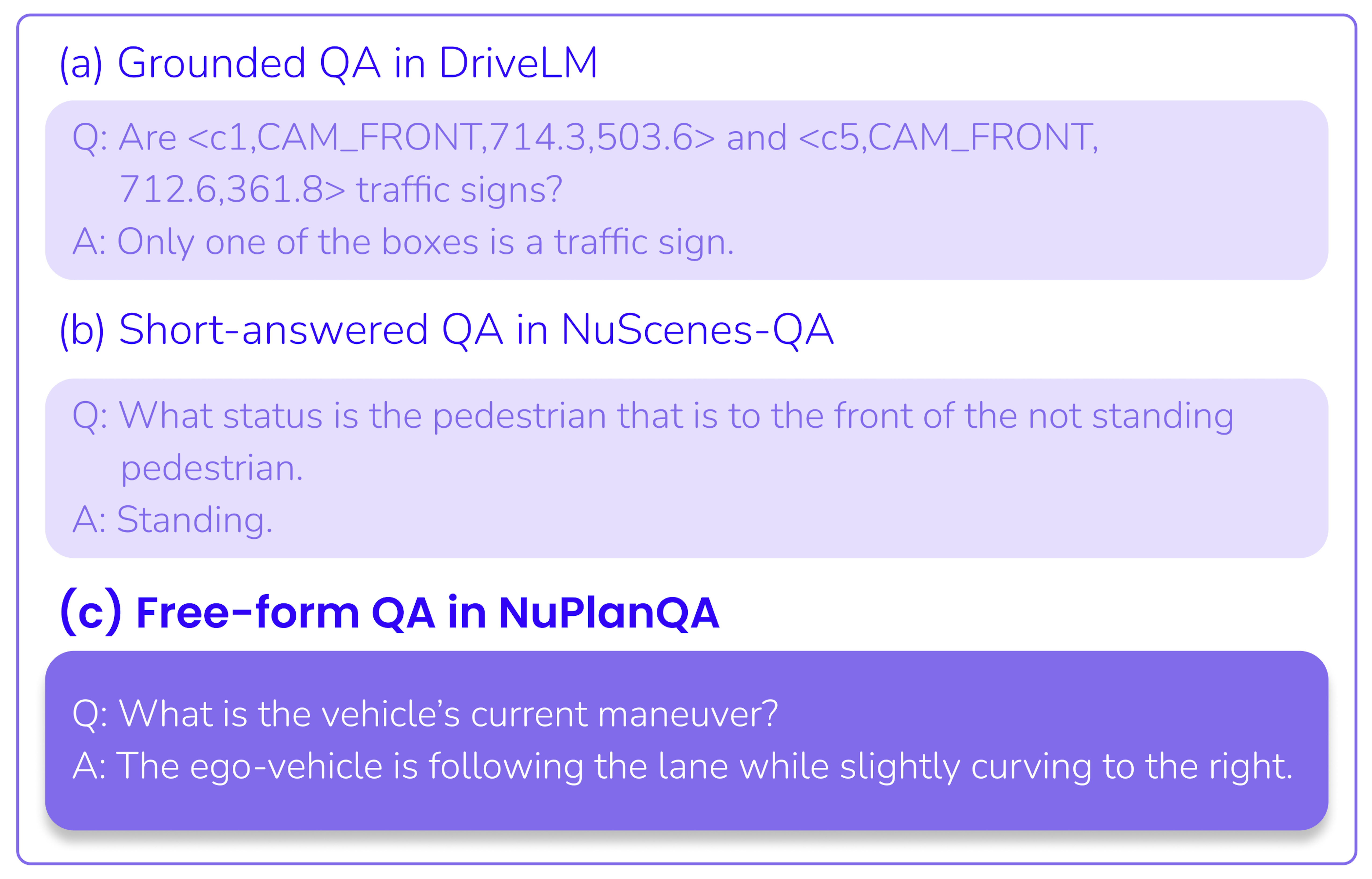}
    \caption{\textbf{Comparison of different QA types.} Free-form QAs offer richer scene descriptions with greater fluency and applicability.}
    \label{fig:qa-task}
\vspace{2mm}
\end{figure}

\subsection{BEV-LLM Framework}

\begin{figure}[t]
    \centering
    \includegraphics[width=\linewidth]{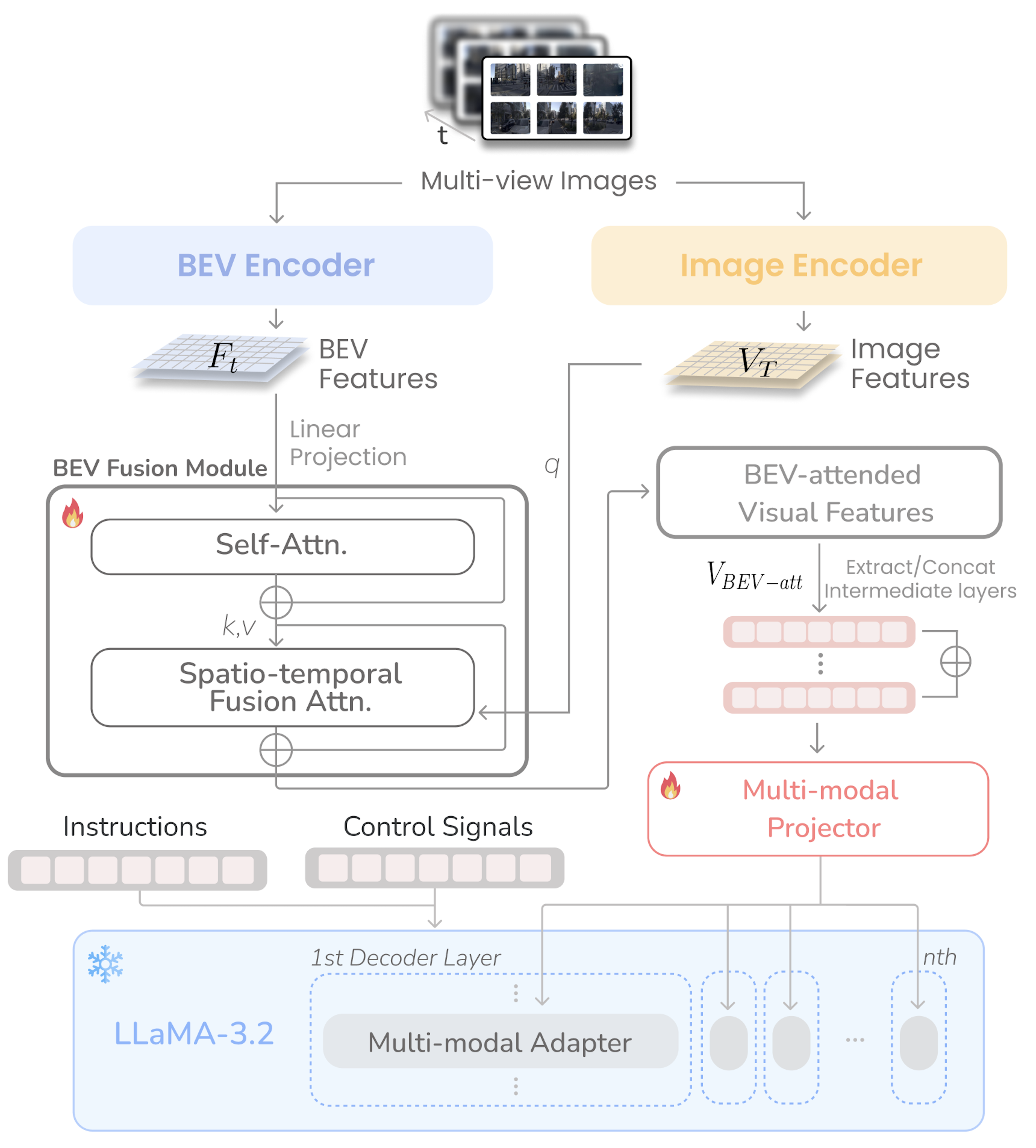}
\caption{\textbf{Architecture of BEV-LLM.} BEV features are integrated into LLaMA via the BEV-Fusion module.}
    \label{fig:model}
\vspace{-3mm}
\end{figure}

In this section, we propose BEV-LLM, which integrates BEV features into MLLMs. Existing MLLM approaches have not sufficiently addressed the efficient integration of images from diverse perspectives. To enhance MLLMs' understanding of multi-view images, we introduce an efficient method for refining image features using BEV representations built upon LLaMA-3.2-Vision-11B \cite{grattafiori2024llama3herdmodels}.

As illustrated in Figure \ref{fig:model}, the time-sequential multi-view images are first processed through the vision encoder of LLaMA-3.2-Vision, which extracts visual features independently for each frame, resulting in per-frame feature representations. The extracted visual features across all timesteps are denoted as \( V_T \), where the visual features extracted at a specific timestep \( t \) are represented as \( V_t \in \mathbb{R}^{B \times N_V \times D_V} \). To ensure that the model captures temporal dependencies across frames, the BEV features are constructed by attending to historical BEV embeddings from previous timesteps using the temporal self-attention (TSA) of BEVFormer \cite{li2022bevformerlearningbirdseyeviewrepresentation} to allow the BEV features at the current timestep to incorporate past information as follows: 

\vspace{-5pt}

\begin{equation}
    F_{\text{t}} = \text{TSA}(Q, \{Q, F_{t-1}\}) + F_{t-1}
\end{equation}

\vspace{3pt}

\noindent where \( F_{t-1} \in \mathbb{R}^{B \times N_F \times D_F} \) are the BEV features from previous timesteps, \( Q \) represents BEV queries interacting with past embeddings, and \( \text{TSA} \) enables self-attention across time, allowing the model to encode motion-aware scene representations. 

Once the BEV features have been refined with historical context, they are fused with the visual features extracted from the multi-view images through spatio-temporal fusion attention (SFA), where the visual features serve as queries, and the temporally-enriched BEV embeddings act as keys and values. This fusion step is formulated as:

\vspace{-5pt}

\begin{equation}
    V_{\text{BEV-att}} = \text{SFA}(V_T, F_{\text{proj}}) + V_T
\end{equation}

\noindent where \( F_{\text{proj}} \) represents the projected motion-aware BEV embeddings from \( F_{\text{t}} \) that contain past temporal information and \( \text{SFA}(V_T, F_{\text{proj}}) \) applies cross-attention, allowing visual features to integrate spatial-temporal BEV context. Since the BEV embeddings already encode motion cues from past frames, this cross-attention inherently transfers temporal awareness to the visual features, making them implicitly aware of past frames even though they are not directly processed with explicit temporal self-attention.

To preserve fine-grained spatial information and enhance alignment, features are extracted from multiple layers of \(V_{\text{BEV-att}}\). Specifically, every fourth layer is extracted along with the final hidden state \cite{grattafiori2024llama3herdmodels}. These intermediate features are then concatenated to form a more robust multi-scale representation. 
To ensure dimensional alignment with the language model, the concatenated representations pass through a multi-modal projector that transforms them into the same embedding space as text features.




At the language model stage, the input text includes natural language instructions and additional information from the vehicle, such as velocity, steering wheel angles, or traffic scenes. Token embeddings from this input text, along with the projected vision-BEV embeddings, are then passed through the decoder layers of LLaMA, which use cross-attention layers in core language model as multi-modal adapter \cite{grattafiori2024llama3herdmodels, alayrac2022flamingovisuallanguagemodel}. Through this interaction, BEV-LLM effectively understands motion, spatial depth, multi-view alignment, and real-time control signals from the vehicle, generating responses that reflect a well-integrated multi-modal understanding.



\setlength{\tabcolsep}{3pt} 

\section{Experiments}

\begin{table*}[!ht]
  \centering
  \small 
\begin{tabularx}{\linewidth}{p{2.7cm} *{13}{>{\centering\arraybackslash}X}}
    \toprule
    \multicolumn{1}{c}{} & \multicolumn{4}{c}{\textbf{Road Env. Perception}} & \multicolumn{4}{c}{\textbf{Spatial Relations Recog.}} & \multicolumn{4}{c}{\textbf{Ego-centric Reasoning}} & \multicolumn{1}{c}{}\\ 
    \cmidrule(lr){2-5} \cmidrule(lr){6-9} \cmidrule(lr){10-13}
    \textbf{\makecell{Method\\ \\}} & \textit{\makecell{Trfc.\\Light}} & \textit{\makecell{Wea\\-ther}} & \textit{\makecell{Road\\Type}} & \textit{\makecell{Avg.}}
    & \textit{\makecell{Sur.\\Obj.}} & \textit{\makecell{Trfc.\\Flow}} & \textit{\makecell{Key\\Obj.}} & \textit{\makecell{Avg.}}
    & \textit{\makecell{Ego\\Ctrl.}} & \textit{\makecell{Situ.\\Asse.}} & \textit{\makecell{Act.\\Rec.}} & \textit{\makecell{Avg.}} & \textbf{\makecell{Total\\ \\}}\\
    \midrule
    \multicolumn{14}{c}{\textbf{Multi-frame as Input}} \\
    \midrule
    GPT-4o {\scriptsize{\cite{openai2024gpt4ocard}}}& 68.5 & 91.7 & 95.0 & 85.1 & 79.6 & 76.5 & 67.7 & 74.6 & 86.8 & 85.1 & 82.4 & 84.8 & 81.5 \\
    Gemini-1.5-Pro {\scriptsize{\cite{geminiteam2024gemini15unlockingmultimodal}}}& 64.5 & 93.5 & 95.8 & 84.6 & 69.1 & 78.6 & 73.2 & 73.6 & 73.8 & 73.8 & 80.4 & 76.0 & 78.1 \\
    \midrule
    VideoLLaMA2{\tiny7B} {\scriptsize{\cite{damonlpsg2024videollama2}}}& \underline{54.2} & 59.4 & \underline{93.8} & 69.1 & 70.7 & \textbf{78.6} & 66.4 & 71.9 & 72.3 & \underline{82.7} & 81.4 & \underline{78.8} & 73.3 \\
    Qwen2.5-VL{\tiny7B} {\scriptsize{\cite{bai2025qwen25vltechnicalreport}}}& 51.7 & 71.9 & 69.3 & 64.3 & 49.2 & 23.0 & 30.9 & 34.4 & 30.4 & 39.6 & 10.1 & 26.7 & 41.8 \\
    LLaVA-OV{\tiny7B} {\scriptsize{\cite{li2024llavaonevisioneasyvisualtask}}}& 53.2 & \textbf{89.4} & \textbf{96.4} & \underline{79.7} & \underline{77.9} & \underline{77.6} & \textbf{73.2} & \textbf{74.3} & \underline{75.4} & 76.2 & \underline{82.9} & 77.5 & \underline{77.8} \\
    LLaVA-NV{\tiny7B} {\scriptsize{\cite{zhang2024llavanext-video}}}& 44.3 & 46.5 & 72.4 & 54.4 & 57.5 & 52.0 & 44.1 & 51.2 & 57.6 & 65.8 & 71.4 & 64.9 & 56.8 \\
    InternVL-1.5{\tiny20B} {\scriptsize{\cite{internvl1.5}}}& 40.9 & 70.5 & 75.5 & 62.3 & 48.6 & 50.0 & 44.5 & 47.7 & 57.1 & 55.0 & 60.8 & 57.6 & 55.9 \\
    LLaVA-NV{\tiny32B} {\scriptsize{\cite{zhang2024llavanext-video}}}& 47.8 & \underline{74.7} & \underline{93.8} & 72.1 & 69.1 & 74.0 & 63.6 & 68.9 & 72.8 & 69.3 & 76.9 & 73.0 & 71.3 \\
    \midrule
    \rowcolor{gray!25} BEV-LLM (Ours)& \textbf{61.1} & \textbf{89.4} & 89.6 & \textbf{80.0} & \textbf{78.5} & 75.5 & \underline{68.2} & \underline{74.1} & \textbf{79.1} & \textbf{83.2} & \textbf{83.4} & \textbf{81.9} & \textbf{78.7} \\
    \midrule
    \multicolumn{14}{c}{\textbf{Single-frame as Input}} \\
    \midrule
    GPT-4o {\scriptsize{\cite{openai2024gpt4ocard}}} & 66.5 & 92.2 & 94.4 & 84.4 & 81.2 & 77.0 & 68.5 & 75.6 & 86.4 & 86.6 & 81.4 & 84.8 & 81.6 \\
    Gemini-1.5-Pro {\scriptsize{\cite{geminiteam2024gemini15unlockingmultimodal}}}& 64.5 & 93.1 & 95.8 & 84.5 & 71.1 & 79.6 & 72.7 & 74.5 & 79.6 & 76.7 & 80.9 & 79.1 & 79.4 \\
    \midrule
    VideoLLaMA2{\tiny7B} {\scriptsize{\cite{damonlpsg2024videollama2}}}& 50.2 & 64.1 & \underline{93.8} & 69.4 & 71.3 & \textbf{79.6} & 64.1 & 71.7 & 69.6 & \textbf{84.2} & \underline{81.4} & \underline{78.4} & 73.2 \\
    Qwen2.5-VL{\tiny7B} {\scriptsize{\cite{bai2025qwen25vltechnicalreport}}}& 51.2 & 70.5 & 62.5 & 61.4 & 47.0 & 16.8 & 23.2 & 29.0 & 27.2 & 40.6 & 12.1 & 26.6 & 39.0 \\
    LLaVA-OV{\tiny7B} {\scriptsize{\cite{li2024llavaonevisioneasyvisualtask}}}& \underline{52.2} & \textbf{93.5} & \textbf{96.4} & \textbf{80.7} & \underline{72.9} & 74.0 & \textbf{68.6} & \underline{71.8} & \underline{72.3} & 72.3 & 79.9 & 74.8 & \underline{75.8} \\
    LLaVA-NV{\tiny7B} {\scriptsize{\cite{zhang2024llavanext-video}}}& 48.8 & 41.5 & 63.5 & 51.3 & 51.9 & 50.5 & 41.8 & 48.1 & 56.5 & 60.4 & 69.8 & 62.2 & 53.9 \\
    InternVL-1.5{\tiny20B} {\scriptsize{\cite{internvl1.5}}}& 39.4 & 68.7 & 72.4 & 60.2 & 54.7 & 48.5 & 41.4 & 48.2 & 62.3 & 58.9 & 58.8 & 60.0 & 56.1 \\
    LLaVA-NV{\tiny32B} {\scriptsize{\cite{zhang2024llavanext-video}}} & 46.8 & 73.3 & 91.7 & 70.6 & 70.7 & 71.9 & 58.2 & 66.9 & 71.2 & 68.8 & \textbf{82.9} & 74.3 & 70.6  \\
    \midrule
    \rowcolor{gray!25} BEV-LLM (Ours)& \textbf{58.1} & \underline{87.1} & 84.9 & \underline{76.7} & \textbf{75.1} & \underline{75.0} & \underline{66.8} & \textbf{72.3} & \textbf{76.0} & \underline{81.2} & \underline{81.4} & \textbf{79.5} & \textbf{76.2} \\
    \bottomrule
  \end{tabularx}
    \vspace{-2mm}
\caption{\textbf{Performance of MLLMs on NuPlanQA-Eval.} The metric used is accuracy (\%)—calculated as the number of correct responses over the total number of questions. The best-performing model in each task is \textbf{bolded}, while the second-best is \underline{underlined}. LLaVA-OV: LLaVA-OneVision, LLaVA-NV: LLaVA-Next-Video.}
  \label{tab:main-results}
\end{table*}

\setlength{\tabcolsep}{7pt} 

\begin{table*}[!ht]
  \small
  \centering
\begin{tabular}{l c c c c c c}
\toprule
\textbf{Method} & \textbf{\makecell{View}} & \textbf{\makecell{Frame}} &\textbf{\makecell{Road Env.\\Perception}} & \textbf{\makecell{Spatial Relation \\Recognition}} & \textbf{\makecell{Ego-centric\\Reasoning}} & \textbf{\makecell{Total}} \\
\midrule
Baseline & \textit{multi-view} & \textit{multi-frame} & 68.7 & 60.8 & 67.0 & 65.5 \\

+BEV-Fusion & \textit{single-view} & \textit{single-frame}& 75.6 \textit{(+6.9)}& 66.0 \textit{(+5.2)}& 72.6 \textit{(+5.6)}& 71.4 \textit{(+5.9)}\\

+BEV-Fusion & \textit{single-view} & \textit{multi-frame}& 75.7 \textit{(+7.0)}& 68.0 \textit{(+7.2)}& 73.6 \textit{(+6.6)}& 72.4 \textit{(+7.4)}\\

+BEV-Fusion & \textit{multi-view} & \textit{single-frame} & \underline{76.7 \textit{(+8.0)}}& \underline{72.3 \textit{(+11.5)}}& \underline{79.5 \textit{(+12.5)}}& \underline{76.2 \textit{(+10.7)}}\\

+BEV-Fusion & \textit{multi-view} & \textit{multi-frame} & \textbf{80.0 \textit{(+11.3)}}& \textbf{74.1 \textit{(+13.3)}}& \textbf{81.9 \textit{(+14.9)}}& \textbf{78.7 \textit{(+13.2)}}\\

\bottomrule
\end{tabular}
\caption{\textbf{Ablations on the BEV-Fusion module and input types.} The baseline is trained on NuPlanQA-1M without BEV-Fusion module. For brevity, performance details for subtasks are provided in the supplementary material. The metric used is accuracy (\%). ``Frame'' indicates whether historical frames are included. Single-view input consists of only the front-view image.}
\label{tab:ablation}
\vspace{-3mm}
\end{table*}


\subsection{Settings}
\noindent \textbf{Training of BEV-LLM.} First, we pretrain the LLaMA-3.2-Vision-11B by activating only the multi-modal projector without incorporating the BEV-Fusion module. Since the original Llama-3.2-Vision is designed for single-timestep images, this pretraining phase aims to adapt the model to handle multi-timestep images. During this phase, we train the model exclusively on front-view images across multiple timesteps. Given our objective of short-term, precise traffic scene understanding, we use the past three frames as input, corresponding to approximately 1.5 seconds of video. After one epoch of training on the NuPlanQA-1M dataset, we integrate the BEV-Fusion module, activating both the multi-modal projector and the BEV-Fusion module, and continue training. 

\vspace{.5em}

\noindent \textbf{Comparison of Recent MLLMs.} We evaluate video LLMs, including both commercial and open-source models, on our NuPlanQA-Eval benchmark to assess their inherent performance in traffic scene analysis. Due to the absence of publicly available multi-view MLLMs incorporating multi-view images, BEV-LLM serves as the multi-view trained baseline. For other MLLMs, we use structured multi-view images as inputs by arranging them into a single image, marking the corresponding camera view for reference. Each model's default settings are used to adjust the frame size accordingly. Since our benchmark consists of multiple-choice questions with four answer options, we adopt accuracy as the evaluation metric, where random guessing yields a baseline accuracy of 25\%.

\subsection{Main Results}
In this section, we compare the performance of MLLMs across nine subtasks using different input types: multi-frame (with historical frames) and single-frame (without historical frames) inputs. As shown in Table \ref{tab:main-results}, when multi-frame inputs are given, BEV-LLM outperforms open-source models in six out of nine subtasks and achieves the highest average score of 78.7\%. It demonstrates particularly strong results in ego-centric reasoning, the skill requiring the highest level of reasoning ability. In this category, BEV-LLM surpasses all other models across all three tasks, outperforming the second-best model by an average of 6.2\%. While its performance for ego-centric reasoning is surpassed by VideoLLaMA2 \cite{damonlpsg2024videollama2} and LLaVA-NV-32B \cite{zhang2024llavanext-video} when using single-frame inputs, BEV-LLM with multi-frame inputs still achieves superior results.


Meanwhile, for road environment perception, VideoLLaMA2 \cite{damonlpsg2024videollama2} , LLaVA-OV-7B \cite{li2024llavaonevisioneasyvisualtask}, and commercial models achieve high scores in weather/lighting conditions and road type/conditions for both multi-frame and single-frame inputs. This is likely because pretrained MLLMs are already well-adapted to these relatively straightforward perception tasks, allowing them to perform strongly. However, traffic light detection proves to be one of the most challenging tasks across models, with seven out of nine models recording their lowest scores in this subtask. Notably, BEV-LLM outperforms other open-source models in this task, achieving 61.1\% accuracy with multi-frame inputs and 58.1\% with single-frame inputs, surpassing the second-best models by 21.7\% and 7.4\%, respectively. This indicates that while existing MLLMs typically perform well at detecting the presence of specific objects, they struggle with interpreting road scenes in a driving context. Specifically, determining which traffic light to prioritize among multiple signals remains a significant challenge.

Among the three core skills, most MLLMs struggle particularly with spatial relations recognition. Even GPT-4o and Gemini-1.5-Pro perform worse in this area, scoring 6.9\% and 4.5\% below their total average, respectively—both lower than in the other two skill areas. One key reason is that road environment perception primarily relies on static visual cues and benefits from pretraining with large-scale image datasets, allowing MLLMs to handle it more easily. Similarly, ego-centric reasoning can be significantly aided by control signals provided as inputs, offering additional context beyond visual information. In contrast, spatial relations recognition depends entirely on visual input and requires temporal reasoning across frames, posing a greater challenge for MLLMs. Due to this fundamental difference, most MLLMs—typically trained on web-scale, long video clips exceeding 1.5 seconds—may be less adept at short-term spatial reasoning in driving scenes. This also suggests that there is substantial room for improvement in this skill for both commercial and open-source MLLMs. To further illustrate these findings, we provide additional qualitative analysis in the supplementary material.



\subsection{Ablation Study}

To further analyze the impact of BEV-Fusion module and input types, we conduct ablation studies. First, to examine the effect of BEV integration on performance, we compare a baseline model—trained on NuPlanQA-1M without the BEV-Fusion module—to BEV-LLM. As shown in Table \ref{tab:ablation}, integrating BEV features through the BEV-Fusion module improves accuracy by 11.3\%, 13.3\%, and 14.9\% for three different skills under the same setting (multi-view, multi-frame). This highlights how BEV representations enhance the capability of multi-view MLLMs by providing a unified perspective and improved spatial awareness. Moreover, the greater enhancements in spatial relations recognition and ego-centric reasoning compared to road environment perception can be attributed to BEV features providing a more stable spatial representation, which remain consistent across frames. In contrast, visual features alone struggle with motion recognition or object tracking due to perspective distortion.

Additionally, we compare results across different input types, including \textit{single- vs. multi-view and single- vs. multi-frame.} When comparing with the baseline, the smallest performance gain is observed with single-view (front-view), single-frame inputs, showing an average improvement of 5.9\% over the baseline. On the other hand, the largest improvement is seen with multi-view, multi-frame inputs, which achieves an average performance boost of 13.2\%. The second-best condition occurs when tested with multi-view, single-frame inputs, highlighting the significant contribution of multi-view representations in both the presence and absence of historical frames. Moreover, results with multi-frame consistently outperform those with single-frame under the same settings, yielding an approximate 5.3\% performance gain. Overall, the ablation study demonstrates that MLLMs can better adapt to driving scenes by effectively integrating BEV features from multi-view images and historical frames, underscoring the potential for extending this approach to other types of MLLMs.

\section{Discussion}
\noindent \textbf{Reasoning on Driving Scenes.} Our evaluation reveals a significant gap between general MLLMs and those designed specifically for driving scenes. Typical MLLMs trained on videos from YouTube, stock footage sites, or other public resources lack high-level recognition skills from a driver's perspective \cite{Bain21, xue2022hdvila}. Moreover, existing MLLMs primarily focus on understanding relatively long video sequences, whereas decision-making for driving requires short-term, instantaneous reasoning within a timeframe of less than 2–3 seconds \cite{zhang2024longcontexttransferlanguage, tan2024koalakeyframeconditionedlong}. Addressing this challenge necessitates denser frame sampling and architectural adjustments for short-term prediction, ensuring more precise decision-making. Additionally, there remains a lack of diverse scenarios in language datasets, particularly those containing various corner cases and complex environments. To improve the generalization of MLLMs across a wider range of driving situations, it is essential to leverage existing datasets that cover diverse scenarios and further develop them into language-oriented datasets.






\vspace{.5em}

\noindent \textbf{Multi-view Understanding of MLLMs.} As revealed in our ablation study, BEV features obtained from multi-view images enhance MLLMs' ability to better perceive surrounding environments. When integrated, BEV features significantly improve performance, particularly in motion-related tasks, aligning with their intended role in autonomous driving. While image features provide rich information on colors and object details, combining them with BEV features enables stable multi-view fusion, allowing models to overcome occlusions and extend their field of view. Additionally, multi-sensor fusion with RADAR and LiDAR is simplified through BEV projection, improving detection robustness. To integrate BEV features into MLLMs, exploring existing techniques for multi-modal understanding—along with appropriate encoder and adapter selection—would be beneficial in facilitating the adoption of various LLMs \cite{holistic_nuinstruct, gao2023llamaadapterv2, blip2, liu2023llava}.


\section{Conclusion}

In this study, we introduced NuPlanQA-Eval, an evaluation benchmark for multi-modal large language models (MLLMs) in driving scene understanding, and NuPlanQA-1M, a large-scale multi-view visual question-answering (VQA) dataset. Our benchmark incorporates nine subtasks across three skills, enabling task-wise comparisons across different MLLMs and facilitating efficient skill enhancement. Our evaluation results indicate that most existing MLLMs struggle with detecting traffic lights and recognizing their colors, as well as understanding spatial relations and ego-centric reasoning, which require contextual analysis of traffic scenes. Additionally, through comparisons with our baseline model, BEV-LLM, we demonstrated that Bird's-Eye-View (BEV) features significantly improve multi-view scene understanding and reasoning in MLLMs, leading to the highest overall score of 78.7\% among open-source models. Although we used LLaMA as the backbone for BEV-LLM, further experiments integrating BEV inputs into various MLLMs could lead to even greater improvements. Exploring alternative approaches to BEV fusion may also enhance model performance. Furthermore, future research can focus on developing reliable foundation models for driving scenes and expanding datasets to cover more diverse scenarios and geographical locations.

{
    \small
    \bibliographystyle{ieeenat_fullname}
    \bibliography{main}

\begin{thebibliography}{55}
\providecommand{\natexlab}[1]{#1}
\providecommand{\url}[1]{\texttt{#1}}
\expandafter\ifx\csname urlstyle\endcsname\relax
  \providecommand{\doi}[1]{doi: #1}\else
  \providecommand{\doi}{doi: \begingroup \urlstyle{rm}\Url}\fi

\bibitem[Alayrac et~al.(2022)Alayrac, Donahue, Luc, Miech, Barr, Hasson, Lenc, Mensch, Millican, Reynolds, Ring, Rutherford, Cabi, Han, Gong, Samangooei, Monteiro, Menick, Borgeaud, Brock, Nematzadeh, Sharifzadeh, Binkowski, Barreira, Vinyals, Zisserman, and Simonyan]{alayrac2022flamingovisuallanguagemodel}
Jean-Baptiste Alayrac, Jeff Donahue, Pauline Luc, Antoine Miech, Iain Barr, Yana Hasson, Karel Lenc, Arthur Mensch, Katie Millican, Malcolm Reynolds, Roman Ring, Eliza Rutherford, Serkan Cabi, Tengda Han, Zhitao Gong, Sina Samangooei, Marianne Monteiro, Jacob Menick, Sebastian Borgeaud, Andrew Brock, Aida Nematzadeh, Sahand Sharifzadeh, Mikolaj Binkowski, Ricardo Barreira, Oriol Vinyals, Andrew Zisserman, and Karen Simonyan.
\newblock Flamingo: a visual language model for few-shot learning, 2022.

\bibitem[Bai et~al.(2025)Bai, Chen, Liu, Wang, Ge, Song, Dang, Wang, Wang, Tang, Zhong, Zhu, Yang, Li, Wan, Wang, Ding, Fu, Xu, Ye, Zhang, Xie, Cheng, Zhang, Yang, Xu, and Lin]{bai2025qwen25vltechnicalreport}
Shuai Bai, Keqin Chen, Xuejing Liu, Jialin Wang, Wenbin Ge, Sibo Song, Kai Dang, Peng Wang, Shijie Wang, Jun Tang, Humen Zhong, Yuanzhi Zhu, Mingkun Yang, Zhaohai Li, Jianqiang Wan, Pengfei Wang, Wei Ding, Zheren Fu, Yiheng Xu, Jiabo Ye, Xi Zhang, Tianbao Xie, Zesen Cheng, Hang Zhang, Zhibo Yang, Haiyang Xu, and Junyang Lin.
\newblock Qwen2.5-vl technical report, 2025.

\bibitem[Bain et~al.(2021)Bain, Nagrani, Varol, and Zisserman]{Bain21}
Max Bain, Arsha Nagrani, G{\"u}l Varol, and Andrew Zisserman.
\newblock Frozen in time: A joint video and image encoder for end-to-end retrieval.
\newblock In \emph{IEEE International Conference on Computer Vision}, 2021.

\bibitem[Banerjee and Lavie(2005)]{banerjee-lavie-2005-meteor}
Satanjeev Banerjee and Alon Lavie.
\newblock {METEOR}: An automatic metric for {MT} evaluation with improved correlation with human judgments.
\newblock In \emph{Proceedings of the {ACL} Workshop on Intrinsic and Extrinsic Evaluation Measures for Machine Translation and/or Summarization}, pages 65--72, Ann Arbor, Michigan, 2005. Association for Computational Linguistics.

\bibitem[Bogdoll et~al.(2021)Bogdoll, Breitenstein, Heidecker, Bieshaar, Sick, Fingscheidt, and Zollner]{corner_cases_Bogdoll_2021}
Daniel Bogdoll, Jasmin Breitenstein, Florian Heidecker, Maarten Bieshaar, Bernhard Sick, Tim Fingscheidt, and J.~Marius Zollner.
\newblock Description of corner cases in automated driving: Goals and challenges.
\newblock In \emph{2021 IEEE/CVF International Conference on Computer Vision Workshops (ICCVW)}. IEEE, 2021.

\bibitem[Caesar et~al.(2020)Caesar, Bankiti, Lang, Vora, Liong, Xu, Krishnan, Pan, Baldan, and Beijbom]{nuscenes_Caesar_2020_CVPR}
Holger Caesar, Varun Bankiti, Alex~H. Lang, Sourabh Vora, Venice~Erin Liong, Qiang Xu, Anush Krishnan, Yu Pan, Giancarlo Baldan, and Oscar Beijbom.
\newblock nuscenes: A multimodal dataset for autonomous driving.
\newblock In \emph{Proceedings of the IEEE/CVF Conference on Computer Vision and Pattern Recognition (CVPR)}, 2020.

\bibitem[Caesar et~al.(2022)Caesar, Kabzan, Tan, Fong, Wolff, Lang, Fletcher, Beijbom, and Omari]{caesar2022nuplanclosedloopmlbasedplanning}
Holger Caesar, Juraj Kabzan, Kok~Seang Tan, Whye~Kit Fong, Eric Wolff, Alex Lang, Luke Fletcher, Oscar Beijbom, and Sammy Omari.
\newblock Nuplan: A closed-loop ml-based planning benchmark for autonomous vehicles, 2022.

\bibitem[Chen et~al.(2024{\natexlab{a}})Chen, han Ding, Wang, Wang, Zhang, and Liu]{chen2024asynchronouslargelanguagemodel}
Yuan Chen, Zi han Ding, Ziqin Wang, Yan Wang, Lijun Zhang, and Si Liu.
\newblock Asynchronous large language model enhanced planner for autonomous driving, 2024{\natexlab{a}}.

\bibitem[Chen et~al.(2024{\natexlab{b}})Chen, Wang, Tian, Ye, Gao, Cui, Tong, Hu, Luo, Ma, Ma, Wang, Dong, Yan, Guo, He, Shi, Jin, Xu, Wang, Wei, Li, Zhang, Zhang, Cai, Wen, Yan, Dou, Lu, Zhu, Lu, Lin, Qiao, Dai, and Wang]{internvl1.5}
Zhe Chen, Weiyun Wang, Hao Tian, Shenglong Ye, Zhangwei Gao, Erfei Cui, Wenwen Tong, Kongzhi Hu, Jiapeng Luo, Zheng Ma, Ji Ma, Jiaqi Wang, Xiaoyi Dong, Hang Yan, Hewei Guo, Conghui He, Botian Shi, Zhenjiang Jin, Chao Xu, Bin Wang, Xingjian Wei, Wei Li, Wenjian Zhang, Bo Zhang, Pinlong Cai, Licheng Wen, Xiangchao Yan, Min Dou, Lewei Lu, Xizhou Zhu, Tong Lu, Dahua Lin, Yu Qiao, Jifeng Dai, and Wenhai Wang.
\newblock How far are we to gpt-4v? closing the gap to commercial multimodal models with open-source suites, 2024{\natexlab{b}}.

\bibitem[Cheng et~al.(2024)Cheng, Leng, Zhang, Xin, Li, Chen, Zhu, Zhang, Luo, Zhao, and Bing]{damonlpsg2024videollama2}
Zesen Cheng, Sicong Leng, Hang Zhang, Yifei Xin, Xin Li, Guanzheng Chen, Yongxin Zhu, Wenqi Zhang, Ziyang Luo, Deli Zhao, and Lidong Bing.
\newblock Videollama 2: Advancing spatial-temporal modeling and audio understanding in video-llms.
\newblock \emph{arXiv preprint arXiv:2406.07476}, 2024.

\bibitem[Cui et~al.(2023)Cui, Ma, Cao, Ye, and Wang]{cui2023drivespeakenablinghumanlike}
Can Cui, Yunsheng Ma, Xu Cao, Wenqian Ye, and Ziran Wang.
\newblock Drive as you speak: Enabling human-like interaction with large language models in autonomous vehicles, 2023.

\bibitem[Ding et~al.(2024)Ding, Han, Xu, Liang, Zhang, and Li]{holistic_nuinstruct}
Xinpeng Ding, Jianhua Han, Hang Xu, Xiaodan Liang, Wei Zhang, and Xiaomeng Li.
\newblock Holistic autonomous driving understanding by bird'view injected multi-modal large models.
\newblock In \emph{2024 IEEE/CVF Conference on Computer Vision and Pattern Recognition (CVPR)}, pages 13668--13677, 2024.

\bibitem[et~al.(2024)]{grattafiori2024llama3herdmodels}
Aaron~Grattafiori et al.
\newblock The llama 3 herd of models, 2024.

\bibitem[Fu et~al.(2024)Fu, Dai, Luo, Li, Ren, Zhang, Wang, Zhou, Shen, Zhang, Chen, Li, Lin, Zhao, Li, Xu, Zheng, Chen, Ji, and Sun]{fu2024videommefirstevercomprehensiveevaluation}
Chaoyou Fu, Yuhan Dai, Yongdong Luo, Lei Li, Shuhuai Ren, Renrui Zhang, Zihan Wang, Chenyu Zhou, Yunhang Shen, Mengdan Zhang, Peixian Chen, Yanwei Li, Shaohui Lin, Sirui Zhao, Ke Li, Tong Xu, Xiawu Zheng, Enhong Chen, Rongrong Ji, and Xing Sun.
\newblock Video-mme: The first-ever comprehensive evaluation benchmark of multi-modal llms in video analysis, 2024.

\bibitem[Gao et~al.(2023)Gao, Han, Zhang, Lin, Geng, Zhou, Zhang, Lu, He, Yue, Li, and Qiao]{gao2023llamaadapterv2}
Peng Gao, Jiaming Han, Renrui Zhang, Ziyi Lin, Shijie Geng, Aojun Zhou, Wei Zhang, Pan Lu, Conghui He, Xiangyu Yue, Hongsheng Li, and Yu Qiao.
\newblock Llama-adapter v2: Parameter-efficient visual instruction model.
\newblock \emph{arXiv preprint arXiv:2304.15010}, 2023.

\bibitem[Gopalkrishnan et~al.(2024)Gopalkrishnan, Greer, and Trivedi]{gopalkrishnan2024multiframelightweightefficient}
Akshay Gopalkrishnan, Ross Greer, and Mohan Trivedi.
\newblock Multi-frame, lightweight \& efficient vision-language models for question answering in autonomous driving, 2024.

\bibitem[Hu et~al.(2023)Hu, Yang, Chen, Li, Sima, Zhu, Chai, Du, Lin, Wang, Lu, Jia, Liu, Dai, Qiao, and Li]{hu2023planningorientedautonomousdriving}
Yihan Hu, Jiazhi Yang, Li Chen, Keyu Li, Chonghao Sima, Xizhou Zhu, Siqi Chai, Senyao Du, Tianwei Lin, Wenhai Wang, Lewei Lu, Xiaosong Jia, Qiang Liu, Jifeng Dai, Yu Qiao, and Hongyang Li.
\newblock Planning-oriented autonomous driving, 2023.

\bibitem[Kim et~al.(2018)Kim, Rohrbach, Darrell, Canny, and Akata]{bddx_kim2018textual}
Jinkyu Kim, Anna Rohrbach, Trevor Darrell, John Canny, and Zeynep Akata.
\newblock Textual explanations for self-driving vehicles.
\newblock \emph{Proceedings of the European Conference on Computer Vision (ECCV)}, 2018.

\bibitem[Li et~al.(2024)Li, Zhang, Guo, Zhang, Li, Zhang, Zhang, Zhang, Li, Liu, and Li]{li2024llavaonevisioneasyvisualtask}
Bo Li, Yuanhan Zhang, Dong Guo, Renrui Zhang, Feng Li, Hao Zhang, Kaichen Zhang, Peiyuan Zhang, Yanwei Li, Ziwei Liu, and Chunyuan Li.
\newblock Llava-onevision: Easy visual task transfer, 2024.

\bibitem[Li et~al.(2023)Li, Li, Savarese, and Hoi]{blip2}
Junnan Li, Dongxu Li, Silvio Savarese, and Steven Hoi.
\newblock Blip-2: bootstrapping language-image pre-training with frozen image encoders and large language models.
\newblock In \emph{Proceedings of the 40th International Conference on Machine Learning}. JMLR.org, 2023.

\bibitem[Li et~al.(2022)Li, Wang, Li, Xie, Sima, Lu, Yu, and Dai]{li2022bevformerlearningbirdseyeviewrepresentation}
Zhiqi Li, Wenhai Wang, Hongyang Li, Enze Xie, Chonghao Sima, Tong Lu, Qiao Yu, and Jifeng Dai.
\newblock Bevformer: Learning bird's-eye-view representation from multi-camera images via spatiotemporal transformers, 2022.

\bibitem[Lin(2004)]{lin-2004-rouge}
Chin-Yew Lin.
\newblock {ROUGE}: A package for automatic evaluation of summaries.
\newblock In \emph{Text Summarization Branches Out}, pages 74--81, Barcelona, Spain, 2004. Association for Computational Linguistics.

\bibitem[Liu et~al.(2023)Liu, Li, Wu, and Lee]{liu2023llava}
Haotian Liu, Chunyuan Li, Qingyang Wu, and Yong~Jae Lee.
\newblock Visual instruction tuning, 2023.

\bibitem[Liu et~al.(2024)Liu, Duan, Zhang, Li, Zhang, Zhao, Yuan, Wang, He, Liu, Chen, and Lin]{mmbench}
Yuan Liu, Haodong Duan, Yuanhan Zhang, Bo Li, Songyang Zhang, Wangbo Zhao, Yike Yuan, Jiaqi Wang, Conghui He, Ziwei Liu, Kai Chen, and Dahua Lin.
\newblock Mmbench: Is your multi-modal model an all-around player?
\newblock In \emph{Computer Vision – ECCV 2024: 18th European Conference, Milan, Italy, September 29–October 4, 2024, Proceedings, Part VI}, page 216–233, Berlin, Heidelberg, 2024. Springer-Verlag.

\bibitem[Ma et~al.(2023)Ma, Cao, Sun, Pavone, and Xiao]{ma2023dolphinsmultimodallanguagemodel}
Yingzi Ma, Yulong Cao, Jiachen Sun, Marco Pavone, and Chaowei Xiao.
\newblock Dolphins: Multimodal language model for driving, 2023.

\bibitem[Ma et~al.(2024)Ma, Cui, Cao, Ye, Liu, Lu, Abdelraouf, Gupta, Han, Bera, Rehg, and Wang]{ma_lampilot_2024}
Yunsheng Ma, Can Cui, Xu Cao, Wenqian Ye, Peiran Liu, Juanwu Lu, Amr Abdelraouf, Rohit Gupta, Kyungtae Han, Aniket Bera, James~M. Rehg, and Ziran Wang.
\newblock {LaMPilot}: {An} {Open} {Benchmark} {Dataset} for {Autonomous} {Driving} with {Language} {Model} {Programs}.
\newblock In \emph{2024 {IEEE}/{CVF} {Conference} on {Computer} {Vision} and {Pattern} {Recognition} ({CVPR})}, pages 15141--15151, Seattle, WA, USA, 2024. IEEE.

\bibitem[Malla et~al.(2022)Malla, Choi, Dwivedi, Choi, and Li]{malla2022dramajointrisklocalization}
Srikanth Malla, Chiho Choi, Isht Dwivedi, Joon~Hee Choi, and Jiachen Li.
\newblock Drama: Joint risk localization and captioning in driving, 2022.

\bibitem[Marcu et~al.(2024)Marcu, Chen, Hünermann, Karnsund, Hanotte, Chidananda, Nair, Badrinarayanan, Kendall, Shotton, Arani, and Sinavski]{marcu2024lingoqavisualquestionanswering}
Ana-Maria Marcu, Long Chen, Jan Hünermann, Alice Karnsund, Benoit Hanotte, Prajwal Chidananda, Saurabh Nair, Vijay Badrinarayanan, Alex Kendall, Jamie Shotton, Elahe Arani, and Oleg Sinavski.
\newblock Lingoqa: Visual question answering for autonomous driving, 2024.

\bibitem[OpenAI et~al.(2024)OpenAI, :, Hurst, and et~al.]{openai2024gpt4ocard}
OpenAI, :, Aaron Hurst, and Adam~Lerer et al.
\newblock Gpt-4o system card, 2024.

\bibitem[Pan et~al.(2024)Pan, Yaman, Nesti, Mallik, Allievi, Velipasalar, and Ren]{VLP_Pan_2024_CVPR}
Chenbin Pan, Burhaneddin Yaman, Tommaso Nesti, Abhirup Mallik, Alessandro~G Allievi, Senem Velipasalar, and Liu Ren.
\newblock Vlp: Vision language planning for autonomous driving.
\newblock In \emph{Proceedings of the IEEE/CVF Conference on Computer Vision and Pattern Recognition (CVPR)}, pages 14760--14769, 2024.

\bibitem[Papineni et~al.(2002)Papineni, Roukos, Ward, and Zhu]{bleu}
Kishore Papineni, Salim Roukos, Todd Ward, and Wei-Jing Zhu.
\newblock Bleu: a method for automatic evaluation of machine translation.
\newblock In \emph{Proceedings of the 40th Annual Meeting on Association for Computational Linguistics}, page 311–318, USA, 2002. Association for Computational Linguistics.

\bibitem[Park et~al.(2024)Park, Lee, Kang, Choi, Park, Cho, Lee, and Kim]{vlaad}
SungYeon Park, MinJae Lee, JiHyuk Kang, Hahyeon Choi, Yoonah Park, Juhwan Cho, Adam Lee, and DongKyu Kim.
\newblock Vlaad: Vision and language assistant for autonomous driving.
\newblock In \emph{2024 IEEE/CVF Winter Conference on Applications of Computer Vision Workshops (WACVW)}, pages 980--987, 2024.

\bibitem[Paul et~al.(2024)Paul, Garg, Choudhary, Singh, and Krishna]{paul2024legodrivelanguageenhancedgoalorientedclosedloop}
Pranjal Paul, Anant Garg, Tushar Choudhary, Arun~Kumar Singh, and K.~Madhava Krishna.
\newblock Lego-drive: Language-enhanced goal-oriented closed-loop end-to-end autonomous driving, 2024.

\bibitem[Pătrăucean et~al.(2023)Pătrăucean, Smaira, Gupta, Continente, Markeeva, Banarse, Koppula, Heyward, Malinowski, Yang, Doersch, Matejovicova, Sulsky, Miech, Frechette, Klimczak, Koster, Zhang, Winkler, Aytar, Osindero, Damen, Zisserman, and Carreira]{pătrăucean2023perceptiontestdiagnosticbenchmark}
Viorica Pătrăucean, Lucas Smaira, Ankush Gupta, Adrià~Recasens Continente, Larisa Markeeva, Dylan Banarse, Skanda Koppula, Joseph Heyward, Mateusz Malinowski, Yi Yang, Carl Doersch, Tatiana Matejovicova, Yury Sulsky, Antoine Miech, Alex Frechette, Hanna Klimczak, Raphael Koster, Junlin Zhang, Stephanie Winkler, Yusuf Aytar, Simon Osindero, Dima Damen, Andrew Zisserman, and João Carreira.
\newblock Perception test: A diagnostic benchmark for multimodal video models, 2023.

\bibitem[Qian et~al.(2025)Qian, Chen, Zhuo, Jiao, and Jiang]{nuscenesqa}
Tianwen Qian, Jingjing Chen, Linhai Zhuo, Yang Jiao, and Yu-Gang Jiang.
\newblock Nuscenes-qa: a multi-modal visual question answering benchmark for autonomous driving scenario.
\newblock In \emph{Proceedings of the Thirty-Eighth AAAI Conference on Artificial Intelligence and Thirty-Sixth Conference on Innovative Applications of Artificial Intelligence and Fourteenth Symposium on Educational Advances in Artificial Intelligence}. AAAI Press, 2025.

\bibitem[Sima et~al.(2024)Sima, Renz, Chitta, Chen, Zhang, Xie, Beißwenger, Luo, Geiger, and Li]{sima2024drivelmdrivinggraphvisual}
Chonghao Sima, Katrin Renz, Kashyap Chitta, Li Chen, Hanxue Zhang, Chengen Xie, Jens Beißwenger, Ping Luo, Andreas Geiger, and Hongyang Li.
\newblock Drivelm: Driving with graph visual question answering, 2024.

\bibitem[Tan et~al.(2024)Tan, Sun, Hu, hsien Wang, Deilamsalehy, Plummer, Russell, and Saenko]{tan2024koalakeyframeconditionedlong}
Reuben Tan, Ximeng Sun, Ping Hu, Jui hsien Wang, Hanieh Deilamsalehy, Bryan~A. Plummer, Bryan Russell, and Kate Saenko.
\newblock Koala: Key frame-conditioned long video-llm, 2024.

\bibitem[Team and et~al.(2024)]{geminiteam2024gemini15unlockingmultimodal}
Gemini Team and Petko~Georgiev et al.
\newblock Gemini 1.5: Unlocking multimodal understanding across millions of tokens of context, 2024.

\bibitem[Wang et~al.(2024)Wang, Yu, Jiang, Lan, Shi, Chang, Kautz, Li, and Alvarez]{wang2024omnidriveholisticllmagentframework}
Shihao Wang, Zhiding Yu, Xiaohui Jiang, Shiyi Lan, Min Shi, Nadine Chang, Jan Kautz, Ying Li, and Jose~M. Alvarez.
\newblock Omnidrive: A holistic llm-agent framework for autonomous driving with 3d perception, reasoning and planning, 2024.

\bibitem[Wei et~al.(2023)Wei, Wang, Schuurmans, Bosma, Ichter, Xia, Chi, Le, and Zhou]{wei2023chainofthoughtpromptingelicitsreasoning}
Jason Wei, Xuezhi Wang, Dale Schuurmans, Maarten Bosma, Brian Ichter, Fei Xia, Ed Chi, Quoc Le, and Denny Zhou.
\newblock Chain-of-thought prompting elicits reasoning in large language models, 2023.

\bibitem[Wei et~al.(2024)Wei, Yuan, Li, Hu, Gan, and Ding]{wei2024occllamaoccupancylanguageactiongenerativeworld}
Julong Wei, Shanshuai Yuan, Pengfei Li, Qingda Hu, Zhongxue Gan, and Wenchao Ding.
\newblock Occllama: An occupancy-language-action generative world model for autonomous driving, 2024.

\bibitem[Wen et~al.(2023)Wen, Yang, Fu, Wang, Cai, Li, Ma, Li, Xu, Shang, Zhu, Sun, Bai, Cai, Dou, Hu, Shi, and Qiao]{wen2023roadgpt4visionearlyexplorations}
Licheng Wen, Xuemeng Yang, Daocheng Fu, Xiaofeng Wang, Pinlong Cai, Xin Li, Tao Ma, Yingxuan Li, Linran Xu, Dengke Shang, Zheng Zhu, Shaoyan Sun, Yeqi Bai, Xinyu Cai, Min Dou, Shuanglu Hu, Botian Shi, and Yu Qiao.
\newblock On the road with gpt-4v(ision): Early explorations of visual-language model on autonomous driving, 2023.

\bibitem[Wu et~al.(2023)Wu, Han, Wang, Liu, Zhang, and Shen]{nuprompt}
Dongming Wu, Wencheng Han, Tiancai Wang, Yingfei Liu, Xiangyu Zhang, and Jianbing Shen.
\newblock Language prompt for autonomous driving.
\newblock \emph{ArXiv}, abs/2309.04379, 2023.

\bibitem[Xu et~al.(2024)Xu, Zhang, Xie, Zhao, Guo, Wong, Li, and Zhao]{xu2024drivegpt4interpretableendtoendautonomous}
Zhenhua Xu, Yujia Zhang, Enze Xie, Zhen Zhao, Yong Guo, Kwan-Yee.~K. Wong, Zhenguo Li, and Hengshuang Zhao.
\newblock Drivegpt4: Interpretable end-to-end autonomous driving via large language model, 2024.

\bibitem[Xue et~al.(2022)Xue, Hang, Zeng, Sun, Liu, Yang, Fu, and Guo]{xue2022hdvila}
Hongwei Xue, Tiankai Hang, Yanhong Zeng, Yuchong Sun, Bei Liu, Huan Yang, Jianlong Fu, and Baining Guo.
\newblock Advancing high-resolution video-language representation with large-scale video transcriptions.
\newblock In \emph{International Conference on Computer Vision and Pattern Recognition (CVPR)}, 2022.

\bibitem[Yang et~al.(2024)Yang, Gao, Qiu, Chen, Li, Dai, Chitta, Wu, Zeng, Luo, Zhang, Geiger, Qiao, and Li]{yang2024genad}
Jiazhi Yang, Shenyuan Gao, Yihang Qiu, Li Chen, Tianyu Li, Bo Dai, Kashyap Chitta, Penghao Wu, Jia Zeng, Ping Luo, Jun Zhang, Andreas Geiger, Yu Qiao, and Hongyang Li.
\newblock {Generalized Predictive Model for Autonomous Driving}.
\newblock In \emph{Proceedings of the IEEE/CVF Conference on Computer Vision and Pattern Recognition (CVPR)}, 2024.

\bibitem[Ye et~al.(2023)Ye, Xu, Xu, Ye, Yan, Zhou, Wang, Hu, Shi, Shi, Li, Xu, Chen, Tian, Qi, Zhang, and Huang]{Ye2023mPLUGOwlME}
Qinghao Ye, Haiyang Xu, Guohai Xu, Jiabo Ye, Ming Yan, Yi Zhou, Junyan Wang, Anwen Hu, Pengcheng Shi, Yaya Shi, Chenliang Li, Yuanhong Xu, Hehong Chen, Junfeng Tian, Qiang Qi, Ji Zhang, and Feiyan Huang.
\newblock mplug-owl: Modularization empowers large language models with multimodality.
\newblock \emph{ArXiv}, abs/2304.14178, 2023.

\bibitem[You et~al.(2024)You, Shi, Jiang, Huang, Gan, Wu, Cheng, Li, and Ran]{you2024v2xvlmendtoendv2xcooperative}
Junwei You, Haotian Shi, Zhuoyu Jiang, Zilin Huang, Rui Gan, Keshu Wu, Xi Cheng, Xiaopeng Li, and Bin Ran.
\newblock V2x-vlm: End-to-end v2x cooperative autonomous driving through large vision-language models, 2024.

\bibitem[Yue et~al.(2024)Yue, Ni, Zhang, Zheng, Liu, Zhang, Stevens, Jiang, Ren, Sun, Wei, Yu, Yuan, Sun, Yin, Zheng, Yang, Liu, Huang, Sun, Su, and Chen]{yue2023mmmu}
Xiang Yue, Yuansheng Ni, Kai Zhang, Tianyu Zheng, Ruoqi Liu, Ge Zhang, Samuel Stevens, Dongfu Jiang, Weiming Ren, Yuxuan Sun, Cong Wei, Botao Yu, Ruibin Yuan, Renliang Sun, Ming Yin, Boyuan Zheng, Zhenzhu Yang, Yibo Liu, Wenhao Huang, Huan Sun, Yu Su, and Wenhu Chen.
\newblock Mmmu: A massive multi-discipline multimodal understanding and reasoning benchmark for expert agi.
\newblock In \emph{Proceedings of CVPR}, 2024.

\bibitem[Zhang et~al.(2024{\natexlab{a}})Zhang, Dai, Lv, and Miao]{zhang2024minidriveefficientvisionlanguagemodels}
Enming Zhang, Xingyuan Dai, Yisheng Lv, and Qinghai Miao.
\newblock Minidrive: More efficient vision-language models with multi-level 2d features as text tokens for autonomous driving, 2024{\natexlab{a}}.

\bibitem[Zhang et~al.(2024{\natexlab{b}})Zhang, Xu, and Li]{chatscene}
Jiawei Zhang, Chejian Xu, and Bo Li.
\newblock Chatscene: Knowledge-enabled safety-critical scenario generation for autonomous vehicles.
\newblock In \emph{2024 IEEE/CVF Conference on Computer Vision and Pattern Recognition (CVPR)}, pages 15459--15469, 2024{\natexlab{b}}.

\bibitem[Zhang et~al.(2024{\natexlab{c}})Zhang, Zhang, Li, Zeng, Yang, Zhang, Wang, Tan, Li, and Liu]{zhang2024longcontexttransferlanguage}
Peiyuan Zhang, Kaichen Zhang, Bo Li, Guangtao Zeng, Jingkang Yang, Yuanhan Zhang, Ziyue Wang, Haoran Tan, Chunyuan Li, and Ziwei Liu.
\newblock Long context transfer from language to vision, 2024{\natexlab{c}}.

\bibitem[Zhang et~al.(2023)Zhang, Carballo, Yang, and Takeda]{Zhang_2023}
Yuxiao Zhang, Alexander Carballo, Hanting Yang, and Kazuya Takeda.
\newblock Perception and sensing for autonomous vehicles under adverse weather conditions: A survey.
\newblock \emph{ISPRS Journal of Photogrammetry and Remote Sensing}, 196:\penalty0 146–177, 2023.

\bibitem[Zhang et~al.(2024{\natexlab{d}})Zhang, Li, Liu, Lee, Gui, Fu, Feng, Liu, and Li]{zhang2024llavanext-video}
Yuanhan Zhang, Bo Li, haotian Liu, Yong~jae Lee, Liangke Gui, Di Fu, Jiashi Feng, Ziwei Liu, and Chunyuan Li.
\newblock Llava-next: A strong zero-shot video understanding model, 2024{\natexlab{d}}.

\bibitem[Zhou et~al.(2024)Zhou, Huang, Bu, Zeng, Li, Qiu, Zhu, Guo, Qiao, and Li]{zhou2024embodiedunderstandingdrivingscenarios}
Yunsong Zhou, Linyan Huang, Qingwen Bu, Jia Zeng, Tianyu Li, Hang Qiu, Hongzi Zhu, Minyi Guo, Yu Qiao, and Hongyang Li.
\newblock Embodied understanding of driving scenarios, 2024.

\end{thebibliography}
}

\end{document}